\documentclass[sigconf]{acmart}

\usepackage{booktabs} 
\usepackage{paralist}

\usepackage[ruled]{algorithm2e} 

\usepackage[colorinlistoftodos]{todonotes} 

\setcopyright{rightsretained} 

\acmConference[]{KDD Bigdas}{August 2017}{Halifax, Canada} 

\acmPrice{15.00}

\begin{document}
\title{A Brief Survey of Text Mining: Classification, Clustering and Extraction Techniques }

\author{Mehdi Allahyari}
\affiliation{%
  \institution{Computer Science Department\\
  University of Georgia\\
Athens, GA}
}
\email{mehdi@uga.edu}

\author{Seyedamin Pouriyeh}
\affiliation{%
  \institution{Computer Science Department\\
  University of Georgia\\
Athens, GA}
}
\email{pouriyeh@uga.edu}

\author{Mehdi Assefi}
  \affiliation{%
  \institution{Computer Science Department\\
  University of Georgia\\
Athens, GA}
}
\email{asf@uga.edu}

\author{Saied Safaei}
\affiliation{%
  \institution{Computer Science Department\\
  University of Georgia\\
Athens, GA}
}
\email{ssa@uga.edu}

\author{Elizabeth D. Trippe}
\affiliation{%
  \institution{Institute of Bioinformatics\\
  University of Georgia\\
Athens, GA}
}
\email{edt37727@uga.edu}

\author{Juan B. Gutierrez}
\affiliation{%
  \institution{Department of Mathematics\\
  Institute of Bioinformatics\\
  University of Georgia\\
Athens, GA}
}
\email{jgutierr@uga.edu}

\author{Krys Kochut}
\affiliation{%
  \institution{Computer Science Department\\
  University of Georgia\\
Athens, GA}
}
\email{kochut@cs.uga.edu}

\renewcommand\shortauthors{Allahyari, M. et al}

\begin{abstract}
The amount of text that is generated every day is increasing dramatically. This tremendous volume of mostly unstructured text cannot be simply processed and perceived by computers. Therefore, efficient and effective techniques and algorithms are required to discover useful patterns. Text mining is the task of extracting meaningful information from text, which has gained significant attentions in recent years. In this paper, we describe several of the most fundamental text mining tasks and techniques including text pre-processing, classification and clustering. Additionally, we briefly explain text mining in biomedical and health care domains. 
\end{abstract}

%
%



\maketitle

\section{Introduction}
Text Mining (TM) field has gained a great deal of attention in recent years due the tremendous amount of text data, which are created in a variety of forms such as social networks, patient records, health care insurance data, news outlets, etc. 
IDC, in a report sponsored by EMC, predicts that the data volume will grow to 40 zettabytes\footnote{\texttt{1 ZB $= 10^{21}$ bytes = 1 billion terabytes.}} by 2020, leading to a 50-time growth from the beginning of 2010 \cite{gantz2012digital}. 

Text data is a good example of unstructured information, which is one of the simplest forms of data that can be generated in most scenarios. Unstructured text is easily processed and perceived by humans, but is significantly harder for machines to understand. Needless to say, this volume of text is an invaluable source of information and knowledge. As a result, there is a desperate need to design methods and algorithms in order to effectively process this avalanche of text in a wide variety of  applications. \\

Text mining approaches are related to traditional data mining, and knowledge discovery methods, with some specificities, as described below. 

\subsection{Knowledge Discovery vs. Data Mining}
There are various definitions for \textbf{\textit{knowledge discovery}} or \textbf{\textit{knowledge discovery in databases (KDD)}} and \textbf{\textit{data mining}} in the literature. We define it as follows: \\
\textbf{Knowledge Discovery in Databases} is extracting implicit valid, new and potentially useful information from data, which is non-trivial \cite{frawley1992knowledge,fayyad1996knowledge}.
\textbf{Data Mining} is a the application of particular algorithms for extracting patterns from data. KDD aims at discovering hidden patterns and connections in the data. Based on the above definitions KDD refers to the overall \emph{process} of discovering useful knowledge from data while \emph{data mining} refers to a specific \emph{step} in this process. Data can be structured like databases, but also unstructured like data in a simple text file.\\

Knowledge discovery in databases is a process that involves several steps to be applied to the data set of interest in order to excerpt useful patterns. These steps are iterative and interactive they may need decisions being made by the user. CRoss Industry Standard Process for Data Mining (Crisp DM\footnote{\texttt{http://www.crisp-dm.org/}}) model defines these primary steps as follows: 
\begin{inparaenum}[\itshape 1\upshape)]
\item understanding of the application and data and identifying the goal of the KDD process,
\item data preparation and preprocessing,
\item modeling,
\item evaluation,
\item deployment.
\end{inparaenum}
Data cleaning and preprocessing is one of the most tedious steps, because it needs special methods to convert textual data to an appropriate format for data mining algorithms to use.

Data mining and knowledge discovery terms are often used interchangeably. Some would consider data mining as \emph{synonym} for knowledge discovery, i.e. data mining consists of all aspects of KDD process. The second definition considers data mining as \emph{part of} the KDD process (see \cite{fayyad1996knowledge}) and explicate the modeling step, i.e. selecting methods and algorithms to be used for searching for patterns in the data. We consider data mining as a modeling phase of KDD process.

Research in knowledge discovery and data mining has seen rapid advances in recent years, because of the vast progresses in hardware and software technology. Data mining continues to evolve from the intersection of diverse fields such as machine learning, databases, statistics and artificial intelligence, to name a few, which shows the underlying interdisciplinary nature of this field. We briefly describe the relations to the three of aforementioned research areas.\\

\emph{Databases} are essential to efficiently analyze large amounts of data. Data mining algorithms on the other hand can significantly boost the ability to analyze the data. Therefore for the data integrity and management considerations, data analysis requires to be integrated with databases \cite{netz2000integration}. An overview for the data mining from the database perspective can be found in \cite{chen1996data}.\\

\emph{Machine Learning} (ML) is a branch of artificial intelligence that tries to define set of approaches to find patterns in data to be able to predict the patterns of future data. Machine learning involves study of methods and algorithms that can extract information automatically. There are a great deal of machine learning algorithms used in data mining. For more information please refer to \cite{mitchell1997machine,sebastiani2002machine}.\\

\emph{Statistics} is a mathematical science that deals with collection, analysis, interpretation or explanation, and presentation of data\footnote{\texttt{http://en.wikipedia.org/wiki/Statistics}}. Today lots of data mining algorithms are based on statistics and probability methods. There has been a tremendous quantities of research for data mining and statistical learning \cite{han2006data,tuffery2011data,aggarwal2012mining,kantardzic2011data}.

\subsection{Text Mining Approaches}
Text Mining or knowledge discovery from text (KDT) $-$ first introduced by Fledman et al. \cite{feldman1995knowledge} $-$ refers to the process of extracting high quality of information from text (i.e. structured such as RDBMS data \cite{dvzeroski2009relational,chen1996data}, semi-structured such as XML and JSON \cite{doroodchi2009investigation,pouriyeh2010secure, pouriyeh2009secure}, and unstructured text resources such as word documents, videos, and images). It widely covers a large set of related topics and algorithms for analyzing text, spanning various communities, including information retrieval, natural language processing, data mining, machine learning many application domains web and biomedical sciences.\\

\textbf{Information Retrieval (IR):} Information Retrieval is the activity of finding information resources (usually documents) from a collection of unstructured data sets that satisfies the information need \cite{manning2008introduction,faloutsos1998survey}. Therefore information retrieval mostly focused on facilitating information access rather than analyzing information and finding hidden patterns, which is the main purpose of text mining. Information retrieval has less priority on processing or transformation of text whereas text mining can be considered as going beyond information access to further aid users to analyze and understand information and ease the decision making.\\

\textbf{Natural Language Processing (NLP):} Natural Language Processing is sub-field of computer science, artificial intelligence and linguistics which aims at understanding of natural language using computers \cite{liddy2001natural,manning1999foundations}. Many of the text mining algorithms extensively make use of NLP techniques, such as part of speech tagging (POG), syntactic parsing and other types of linguistic analysis (see \cite{rajman1998text,kao2007natural} for more information).\\

\textbf{Information Extraction from text (IE):} Information Extraction is the task of automatically extracting information or facts from unstructured or semi-structured documents \cite{cowie1996information,sarawagi2008information}. It usually serves as a starting point for other text mining algorithms. For example extraction entities, Name Entity Recognition (NER), and their relations from text can give us useful semantic information.\\  

\textbf{Text Summarization:} Many text mining applications need to summarize the text documents in order to get a concise overview of a large document or a collection of documents on a topic \cite{radev2002introduction,hotho2005brief}. There are two categories of summarization techniques in general: \emph{extractive} summarization where a summary comprises information units extracted from the original text, and in contrary \emph{abstractive} summarization where a summary may contain ``synthesized'' information that may not occur in the original document (see \cite{das2007survey, arXiv170702268A} for an overview).\\

\textbf{Unsupervised Learning Methods:} Unsupervised learning methods are techniques trying to find hidden structure out of unlabeled data. They do not need any training phase, therefore can be applied to any text data without manual effort. \emph{Clustering} and \emph{topic modeling} are the two commonly used unsupervised learning algorithms used in the context of text data. Clustering is the task of segmenting a collection of documents into partitions where documents in the same group (cluster) are more similar to each other than those in other clusters. In topic modeling a probabilistic model is used to determine a \emph{soft} clustering, in which every document has a probability distribution over all the clusters as opposed to hard clustering of documents. In topic models each \emph{topic} can be represented as a probability distributions over words and each documents is expressed as probability distribution over topics. Thus, a topic is akin to a cluster and the membership of a document to a topic is probabilistic \cite{aggarwal2012mining,steyvers2007probabilistic}.\\

\textbf{Supervised Learning Methods:} Supervised learning methods are machine learning techniques pertaining to infer a function or learn a classifier from the training data in order to perform predictions on unseen data. There is a broad range of supervised methods such as nearest neighbor classifiers, decision trees, rule-based classifiers and probabilistic classifiers \cite{sebastiani2002machine,mitchell1997machine}.\\

\textbf{Probabilistic Methods for Text Mining:} There are various probabilistic techniques including unsupervised topic models such as probabilistic Latent semantic analysis (pLSA) \cite{hofmann1999probabilistic} and Latent Dirichlet Allocation (LDA) \cite{blei2003latent}, and supervised learning methods such as conditional random fields \cite{lafferty2001conditional} that can be used regularly in the context of text mining. \\

\textbf{Text Streams and Social Media Mining:} There are many different applications on the web which generate tremendous amount of streams of text data. news stream applications and aggregators such as Reuters and Google news generate huge amount of text streams which provides an invaluable source of information to mine. Social networks, particularly Facebook and Twitter create large volumes of text data continuously. They provide a platform that allows users to freely express themselves in a wide range of topics. The dynamic nature of social networks makes the process of text mining difficult which needs special ability to handle poor and non-standard language \cite{gundecha2012mining,yang2012social}.\\

\textbf{Opinion Mining and Sentiment Analysis:} With the advent of e-commerce and online shopping, a huge amount of text is created and continues to grow about different product reviews or users opinions. By mining such data we find important information and opinion about a topic which is significantly fundamental in advertising and online marketing (see \cite{pang2008opinion} for an overview).\\

\textbf{Biomedical Text Mining:} Biomedical text mining refers to the task of text mining on text of biomedical sciences domains. The role of text mining in biomedical domain is two fold, it enables the biomedical researchers to efficiently and effectively access and extract the knowledge out of the massive volumes of data and also facilitates and boosts up biomedical discovery by augmenting the mining of other biomedical data such as genome sequences and protein structures \cite{gutierrez2015within}.\\

\section{Text Representation and Encoding} Text mining on a large collection of documents is usually a complex process, thus it is critical to have a data structure for the text which facilitates further analysis of the documents \cite{hotho2005brief}. The most common way to represent the documents is as a\emph{bag of words} (BOW), which considers the number of occurrences of each term (word/phrase) but ignores the order. This representation leads to a vector representation that can be analyzed with dimension reduction algorithms from machine learning and statistics. Three of the main dimension reduction techniques used in text mining are \emph{Latent Semantic Indexing} (LSI) \cite{dumais1995latent}, \emph{Probabilistic Latent Semantic Indexing} (PLSA) \cite{hofmann1999probabilistic} and \emph{topic models} \cite{blei2003latent}.\\

In many text mining applications, particularly information retrieval (IR), documents needs to be ranked for more effective retrieval over large collections \cite{singhal2001modern}. In order to be able to define the importance of a word in a document, documents are represented as vectors and a numerical \emph{importance} is assigned to each word. The three most used model based on this idea are vector space model (VSM) (see section 2.2), probabilistic models \cite{manning2008introduction} and inference network model \cite{manning2008introduction,turtle1989inference}.\\

\subsection{Text Preprocessing}
\emph{Preprocessing} is one of the key components in many text mining algorithms. For example a traditional text categorization framework comprises preprocessing, feature extraction, feature selection and classification steps. Although it is confirmed that feature extraction \cite{gunal2006feature}, feature selection \cite{feng2012bayesian} and classification algorithm \cite{tan2011adapting} have significant impact on the classification process, the preprocessing stage may have noticeable influence on this success. Uysal et al. \cite{uysal2014impact} have investigated the impact of preprocessing tasks particularly in the area of text classification. The preprocessing step usually consists of the tasks such as tokenization, filtering, lemmatization and stemming. In the following we briefly describe them.\\

\emph{\textbf{Tokenization:}} Tokenization is the task of breaking a character sequence up into pieces (words/phrases) called tokens, and perhaps at the same time throw away certain characters such as punctuation marks. The list of tokens then is used to further processing \cite{webster1992tokenization}.\\

\emph{\textbf{Filtering:}} Filtering is usually done on documents to remove some of the words. A common filtering is stop-words removal. Stop words are the words frequently appear in the text without having much content information (e.g. prepositions, conjunctions, etc). Similarly words occurring quite often in the text said to have little information to distinguish different documents and also words occurring very rarely are also possibly of no significant relevance and can be removed from the documents \cite{saif2014stopwords,silva2003importance}.\\

\emph{\textbf{Lemmatization:}} Lemmatization is the task that considers the morphological analysis of the words, i.e. grouping together the various inflected forms of a word so they can be analyzed as a single item. In other words lemmatization methods try to map verb forms to infinite tense and nouns to a single form. In order to lemmatize the documents we first must specify the POS of each word of the documents and because POS is tedious and error prone, in practice \emph{stemming} methods are preferred.\\

\emph{\textbf{Stemming:}} Stemming methods aim at obtaining stem (root) of derived words. Stemming algorithms are indeed language dependent. The first stemming algorithm introduced in \cite{lovins1968development}, but the stemmer published in \cite{porter1980algorithm} is most widely stemming method used in English \cite{hull1996stemming}.\\

\subsection{Vector Space Model}
In order to allow for more formal descriptions of the algorithms, we first define some terms and variables that will be frequently used in the following: Given a collection of documents $\mathcal{D}=\{d_1, d_2, \ldots, d_D\}$, let $\mathcal{V}=\{w_1, w_2, \ldots, w_v\}$ be the set of distinct words/terms in the collection. Then $\mathcal{V}$ is called the \emph{vocabulary}. The \emph{frequency} of the term $w$ $\in \mathcal{V}$ in document $d\in \mathcal{D}$ is shown by $f_d(w)$ and the number of documents having the word $w$ is represented by $f_\mathcal{D}(w)$ . The term vector for document $d$ is denoted by $\vec{t_d}=(f_d(w_1), f_d(w_2), \ldots,\\ f_d(w_v))$.\\

The most common way to represent documents is to convert them into numeric vectors. This representation is called ``Vector Space Model'' (VSM). Event though its structure is simple and originally introduced for indexing and information retrieval \cite{salton1975vector}, VSM is broadly used in various text mining algorithms and IR systems and enables efficient analysis of large collection of documents \cite{hotho2005brief}.\\

In VSM each word is represented by a variable having a numeric value indicating the \emph{weight} (importance) of the word in the document. There are two main term weight models:
\begin{inparaenum}[ \itshape 1\upshape)]
\item \emph{Boolean model}: In this model a weight $\omega_{ij} > 0$ is assigned to each term $w_i\in d_j$. For any term that does not appear in $d_j, \omega_{ij} = 0$.
\item \emph{Term frequency-inverse document frequency} (TF-IDF): The most popular term weighting schemes is TF-IDF. Let $q$ be this term weighting scheme, then the weight of each word $w \in d$ is computed as follows:\\
\end{inparaenum}

\begin{equation}
	q(w) = f_d(w) * log \dfrac{|\mathcal{D}|}{f_\mathcal{D}(w)}
\end{equation} 
where  $|\mathcal{D}|$ is the number of documents in the collection $\mathcal{D}$.

 In TF-IDF the term frequency is normalized by \emph{inverse document frequency}, IDF. This normalization decreases the weight of the terms occurring more frequently in the document collection, Making sure that the matching of documents be more effected by distinctive words which have relatively low frequencies in the collection.\\

Based on the term weighting scheme, each document is represented by a vector of term weights $\omega(d) = (\omega(d,w_1), \omega(d,w_2), \ldots,\\ \omega(d,w_v))$. We can compute the similarity between two documents $d_1$ and $d_2$. One of the most widely used similarity measures is cosine similarity and is computed as follows:\\
\begin{equation}
	S(d_1, d_2) = cos(\theta) = \dfrac{d_1 \cdot d_2}{\sqrt{\sum\limits_{i = 1}^v{w_{1i}^2}} \cdot \sqrt{\sum\limits_{i = 1}^v{w_{2i}^2}}}
\end{equation}

\cite{salton1988term,salton1975vector} discussed term weighting schemes and vector space models in more details.

\section{Classification}
Text classification has been broadly studied in different communities such as data mining, database, machine learning and information retrieval, and used in vast number of applications in various domains such as image processing, medical diagnosis, document organization, etc. Text classification aims to assign predefined classes to text documents \cite{mitchell1997machine}. The problem of classification is defined as follows. We have a \emph{training set} $\mathcal{D} = \{d_1, d_2, \ldots, d_n\}$ of documents, such that each document $d_i$ is labeled with a label $\ell_i$ from the set $\mathcal{L} = \{\ell_1, \ell_2, \ldots, \ell_k\}$. The task is to find a \emph{classification model} (classifier) $f$ where \\
\begin{equation}
	f : \mathcal{D} \longrightarrow \mathcal{L}	 \hspace{50pt} f(d) = \ell
\end{equation}
which can assign the correct class label to new document $d$ (test instance). The classification is called \emph{hard}, if a label is explicitly assigned to the test instance and \emph{soft}, if a probability value is assigned to the test instance. There are other types of classification which allow assignment of multiple labels \cite{gopal2010multilabel} to a test instance. For an extensive overview on a number of classification methods see \cite{duda2012pattern,james1985classification}. Yang et al.  evaluates various kinds of text classification algorithms \cite{yang1999re}. Many of the classification algorithms have been implemented in different software systems and are publicly available such as BOW toolkit \cite{mccallum1996bow}, Mallet \cite{mccallum2002mallet} and WEKA\footnote{\texttt{http://www.cs.waikato.ac.nz/ml/weka/}}.

To evaluate the performance of the classification model, we set a side a random fraction of the labeled documents (test set). After training the classifier with training set, we classify the test set and compare the estimated labels with the true labels and measure the performance. The portion of correctly classified documents to the total number of documents is called \emph{accuracy} \cite{hotho2005brief}. The common evaluation metrics for text classification are precision, recall and F-1 scores. Charu et al. \cite{aggarwal2012mining} defines the metrics as follows: ``\emph{precision} is the fraction of the correct instances among the identified positive instances. \emph{Recall} is the percentage of correct instances among all the positive instances. And \emph{F-1 score} is the geometric mean of precision and recall''.
\begin{equation}
	F_1 = 2 \times \dfrac{precision \times recall}{precision + recall}
\end{equation}

\subsection{Naive Bayes Classifier} Probabilistic classifiers have gained a lot of popularity recently and have shown to perform remarkably well \cite{chakrabarti1997using,joachims1996probabilistic,koller1997hierarchically,larkey1996combining,sahami1998bayesian}. These probabilistic approaches make assumptions about how the data (words in documents) are generated and propose a probabilistic model based on these assumptions. Then use a set of training examples to estimate the parameters of the model. Bayes rule is used to classify new examples and select the class that is most likely has generated the example \cite{mccallum1998comparison}.\\
The \emph{Naive Bayes} classifier is perhaps the simplest and the most widely used classifier. It models the distribution of documents in each class using a probabilistic model assuming that the distribution of different terms are \emph{independent} from each other. Even though this so called ``naive Bayes'' assumption is clearly false in many real world applications, naive Bayes performs surprisingly well.

There are two main models commonly used for naive Bayes classifications \cite{mccallum1998comparison}. Both models aim at finding the posterior probability of a class, based on the distribution of the words in the document. They difference between these two models is, one model takes into account the frequency of the words whereas the other one does not.

\begin{enumerate}
	\item \textbf{Multi-variate Bernoulli Model:} In this model a document is represented by a vector of binary features denoting the presence or absence of the words in the document. Thus, the frequency of words are ignored. The original work can be found in \cite{lewis1998naive}.

	\item \textbf{Multinomial Model:} We capture the frequencies of words (terms) in a document by representing the document as bag of words. Many different variations of multinomial model have been introduced in  \cite{kalt1996new,mccallum1998improving,mitchell1997machine,nigam1998learning}. McCallum et al. \cite{mccallum1998comparison} have done an extensive comparison between Bernoulli and multinomial models and concluded that 
	\begin{itemize}
		\item If the size of the vocabulary is small, the Bernoulli model may outperform multinomial model.
		\item The multinomial model always outperforms Bernoulli model for large vocabulary sizes, and almost always performs better than Bernoulli if the size of the vocabulary chosen optimally for both models.
	\end{itemize}
\end{enumerate}

Both of these models assume that the documents are generated by a mixture model parameterized by $\theta$. We use the framework McCallum el at. \cite{mccallum1998comparison} defined as follows:\\

The mixture model comprises mixture components $c_j \in \mathcal{C} = \{c_1, c_2, \ldots, c_k\}$. Each document $d_i = \{w_1, w_2, \ldots, w_{n_{i}}\}$ is generated by first selecting a component according to priors, $P(c_j|\theta)$ and then use the component to create the document according to its own parameters, $P(d_i|c_j;\theta)$. Hence, we can compute the likelihood of a document using the sum of probabilities over all mixture components:
\begin{equation}
	P(d_i|\theta) = \sum\limits_{j = 1}^k{P(c_j|\theta)P(d_i|c_j;\theta)}
\end{equation}

We assume a one to one correspondence between classes $\mathcal{L} = \{\ell_1, \ell_2, \ldots, \ell_k\}$ and mixture components, and therefore $c_j$ indicates both the $j$th mixture component and the $j$th class. Consequently, Given a set of labeled training examples, $\mathcal{D} = \{d_1, d_2, \ldots, d_{|\mathcal{D}|}\}$, we first learn (estimate) the parameters of the probabilistic classification model, $\hat{\theta}$, and then using the estimates of these parameters, we perform the classification of test documents by calculating the posterior probabilities of each class $c_j$, given the test document, and select the most likely class (class with the highest probability).
\begin{equation}
	\begin{split}
		P(c_j|d_i;\hat{\theta}) & = \dfrac{P(c_j|\hat{\theta}) P(d_i|c_j;\hat{\theta}_j)} {P(d_i|\hat{\theta})} \\
		& = \dfrac{P(c_j|\hat{\theta}) P(w_1, w_2, \ldots, w_{n_{i}}|c_j;\hat{\theta}_j)} {\sum\limits_{c \in \mathcal{C}}P(w_1, w_2, \ldots, w_{n_{i}}|c;\hat{\theta}_c) P(c|\hat{\theta})} 
	\end{split}
\end{equation}

where based on naive Bayes assumption, words in a document are independent of each other, thus:
\begin{equation}
	P(w_1, w_2, \ldots, w_{n_{i}}|c_j;\hat{\theta}_j) = \displaystyle\prod\limits_{i=1}^{n_i}{P(w_i|c_j;\hat{\theta}_j)}
\end{equation}

\subsection{Nearest Neighbor Classifier}
Nearest neighbor classifier is a proximity-based classifier which use distance-based measures to perform the classification. The main idea is that documents which belong to the same class are more likely ``similar'' or close to each other based on the similarity measures such as cosine defined in (2.2). The classification of the test document is inferred from the class labels of the similar documents in the training set. If we consider the $k$-nearest neighbor in the training data set, the approach is called \emph{$k$-nearest neighbor classification} and the most common class from these $k$ neighbors is reported as the class label, see \cite{manning2008introduction,sebastiani2002machine,han2001text, rezaeiye2014use} for more information and examples.

\subsection{Decision Tree classifiers}
Decision tree is basically a hierarchical tree of the training instances, in which a condition on the attribute value is used to divide the data hierarchically. In other words decision tree \cite{friedl1997decision} recursively partitions the training data set into smaller subdivisions based on a set of tests defined at each node or branch. Each node of the tree is a test of some \emph{attribute} of the traning instance, and each branch descending from the node corresponds to one the value of this attribute. An instance is classified by beginning at the root node, testing the attribute by this node and moving down the tree branch corresponding to the value of the attribute in the given instance. And this process is recursively repeated \cite{mitchell1997machine}.

In case of text data, the conditions on the decision tree nodes are commonly defined in terms of terms in the text documents. For instance a node may be subdivided to its children relying on the presence or absence of a particular term in the document. For a detailed discussion of decision trees see \cite{breiman1984classification,duda2012pattern,james1985classification,quinlan1986induction}.

Decision trees have been used in combination with boosting techniques. \cite{freund1997decision,schapire2000boostexter} discuss boosting techniques to improve the accuracy of the decision tree classification. 

\subsection{Support Vector Machines}
Support Vector Machines (SVM) are supervised learning classification algorithms where have been extensively used in text classification problems. SVM are a form of \emph{Linear Classifiers}. Linear classifiers in the context of text documents are models that making a classification decision is based on the value of the linear combinations of the documents features. Thus, the output of a linear predictor is defined to be $y = \vec{a}\cdot \vec{x} + b$, where $\vec{x} = (x_1,x_2, \ldots, x_n)$ is the normalized document word frequency vector, $\vec{a} = (a_1,a_2, \ldots, a_n)$ is vector of coefficients and $b$ is a scalar. We can interpret the predictor $y = \vec{a}\cdot \vec{x} + b$ in the categorical class labels as a \emph{separating hyperplane} between different classes. 

The SVM initially introduced in \cite{Vapnik:1982:EDB:1098680,cortes1995support}. Support Vector Machines try to find a ``good'' linear separators between various classes \cite{cortes1995support,vapnik2000nature}. A single SVM can only separate two classes, a positive class and a negative class \cite{hotho2005brief}. SVM algorithm attempts to find a hyperplane with the maximum distance $\xi$ (also called \emph{margin}) from the positive and negative examples. The documents with distance $\xi$ from the hyperplane are called \emph{support vectors} and specify the actual location of the hyperplane. If the document vectors of the two classes are not linearly separable, a hyperplane is determined such that the least number of document vectors are located in the wrong side. 

One advantage of the SVM method is that, it is quite robust to high dimensionality, i.e. learning is almost independent of the dimensionality of the feature space. It rarely needs feature selection since it selects data points (support vectors) required for the classification \cite{hotho2005brief}. Joachims et al. \cite{joachims1998text} has described that text data is an ideal choice for SVM classification due to sparse high dimensional nature of the text with few irrelevant features. SVM methods have been widely used in many application domains such as pattern recognition, face detection and spam filtering \cite{burges1998tutorial,osuna1997training,drucker1999support}. For a deeper theoretical study of SVM method see \cite{joachims2001statistical}.

\section{Clustering}
Clustering is one of the most popular data mining algorithms and have extensively studied in the context of text. It has a wide range of applications such as in classification \cite{baker1998distributional,bekkerman2001feature}, visualization \cite{cadez2003model} and document organization \cite{cutting1992scatter}. The clustering is the task of finding groups of similar documents in a collection of documents. The similarity is computed by using a similarity function. Text clustering can be in different levels of granularities where clusters can be documents, paragraphs, sentences or terms. Clustering is one of the main techniques used for organizing documents to enhance retrieval and support browsing, for example Cutting et al. \cite{cutting1993constant} have used clustering to produce a table of contents of a large collection of documents. \cite{anick1997exploiting} exploits clustering to construct a context-based retrieval systems. For a broad overview of clustering see \cite{jain1988algorithms,kaufman2009finding}. There are various software tools such as Lemur\footnote{\texttt{http://www.lemurproject.org/}} and BOW \cite{mccallum1996bow} which have implementations of common clustering algorithms.\\

There are many clustering algorithms that can be used in the context of text data. Text document can be represented as a binary vector, i.e. considering the presence or absence of word in the document. Or we can use more refined representations which involves weighting methods such as TF-IDF (see section 2.2). 

Nevertheless, such naive methods do not usually work well for text clustering, since text data has a number of distinct characteristics which demands the design of text-specific algorithms for the task. We describe some of these unique properties of text representation:
\begin{itemize}
	\item [i.] Text representation has a very large dimensionality, but the underlying data is sparse. In other words, the size of the vocabulary from which the documents are drawn is massive (e.g. order of $10^5$), but a given document may have only a few hundred words. This problem becomes even more severe when we deal with short data such as tweets.

	\item [ii.] Words of the vocabulary of a given collection of documents are commonly correlated with each other. i.e. the number of concepts in the data are much smaller that the feature space. Thus, we need to design algorithms which take the word correlation into consideration in the clustering task.

	\item [iii.] Since documents differ from one another in terms of the number of words they contain, normalizing document representations during the clustering process is important.
\end{itemize}

The aforementioned text characteristics necessitates the design of specialized algorithms for representing text and broadly investigated in IR community and many algorithms have been proposed to optimize text representation \cite{salton1988term}.\\

Text clustering algorithms are split into many different types such as agglomerative clustering algorithms, partitioning algorithms and probabilistic clustering algorithms. Clustering algorithms have varied trade offs in terms of effectiveness and efficiency. For an experimental comparison of different clustering algorithms see \cite{steinbach2000comparison,zhao2004empirical}, and for a survey of clustering algorithms see \cite{xu2005survey}. In the following we describe some of most common text clustering algorithms.

\subsection{Hierarchical Clustering algorithms}
Hierarchical clustering algorithms received their name because they build a group of clusters that can be depicted as a hierarchy of clusters. The hierarchy can be constructed in top-down (called \emph{divisive}) or bottom-up (called \emph{agglomerative}) fashion. Hierarchical clustering algorithms are one of the \emph{Distanced-based clustering algorithms}, i.e. using a similarity function to measure the closeness between text documents. An extensive overview of the hierarchical clustering algorithms for text data is found in \cite{murtagh1983survey,murtagh1984complexities,willett1988recent}.

In the top-down approach we begin with one cluster which includes all the documents. we recursively split this cluster into sub-clusters. In the agglomerative approach, each document is initially considered as an individual cluster. Then successively the most similar clusters are merged together until all documents are embraced in one cluster. There are three different merging methods for agglomerative algorithms:
\begin{inparaenum}[\itshape 1\upshape)]
	\item \emph{Single Linkage Clustering:} In this technique, the similarity between two groups of documents is the highest similarity between any pair of documents from these groups.
	\item \emph{Group-Average Linkage Clustering:} In group-average clustering, the similarity between two cluster is the \emph{average} similarity between pairs of documents in these groups.
	\item \emph{Complete Linkage Clustering:} In this method, the similarity between two clusters is the \emph{worst case} similarity between any pair of documents in these groups. For more information about these merging techniques see \cite{aggarwal2012mining}.
\end{inparaenum}

\subsection{\textit{k}-means Clustering}
$k$-means clustering is one the \emph{partitioning} algorithms which is widely used in the data mining. The $k$-means clustering, partitions $n$ documents in the context of text data into $k$ clusters. representative around which the clusters are built. The basic form of $k$-means algorithm is:\\

\begin{algorithm}
\SetKwInOut{Input}{Input}\SetKwInOut{Output}{Output}
\Input{Document set $\mathcal{D}$, similarity measure \emph{$\mathcal{S}$}, number $k$ of cluster}
\Output{Set of $k$ clusters}
\DontPrintSemicolon
\emph{initialization}\;
Select randomly $k$ data points as starting centroids.\;
\While{not converged}{
		Assign documents to the centroids based on the closest similarity. \;
		Calculate the the cluster centroids for all the clusters.\;
	}
\KwRet{$k$ clusters}
\caption{$k$-means clustering algorithm}
\end{algorithm}

~\\
Finding an optimal solution for $k$-means clustering is computationally difficult (NP-hard), however, there are efficient heuristics such as \cite{bradley1998refining} that are employed in order to converge rapidly to a local optimum. The main disadvantage of $k$-means clustering is that it is indeed very sensitive to the initial choice of the number of $k$. Thus, there are some techniques used to determine the initial $k$, e.g. using another lightweight clustering algorithm such as agglomerative clustering algorithm. More efficient $k$-means clustering algorithms can be found in \cite{kanungo2002efficient,alsabti1997efficient}.

\subsection{Probabilistic Clustering and Topic Models}
\emph{Topic modeling} is one of the most popular the probabilistic clustering algorithms which has gained increasing attention recently. The main idea of topic modeling \cite{blei2003latent,griffiths2002probabilistic,hofmann1999probabilistic} is to create a \emph{probabilistic generative model} for the corpus of text documents. In topic models, documents are mixture of topics, where a topic is a probability distribution over words.

The two main topic models are \emph{Probabilistic Latent Semantic Analysis (pLSA)} \cite{hofmann1999probabilistic} and \emph{Latent Dirichlet Allocation (LDA)} \cite{blei2003latent}. Hofmann (1999) introduced pLSA for document modeling. pLSA model does not provide any probabilistic model at the document level which makes it difficult to generalize it to model new unseen documents. Blei et al. \cite{blei2003latent} extended this model by introducing a Dirichlet prior on mixture weights of topics per documents, and called the model Latent Dirichlet Allocation (LDA). In this section we describe the LDA method. 

\begin{figure}[t]
	\centering
	\includegraphics[scale=0.6]{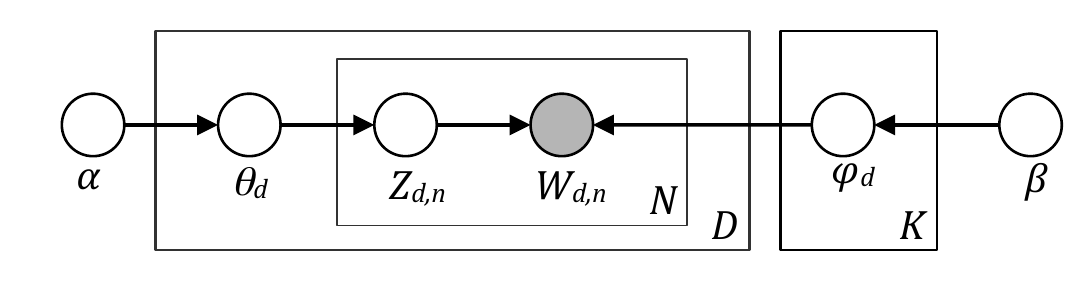}
\caption{LDA Graphical Model}
\label{Fig:LDA}
\end{figure}

The latent Dirichlet allocation model is the state of the art unsupervised technique for extracting thematic information (topics) of a collection of documents. \cite{blei2003latent,griffiths2004finding}. The basic idea is that documents are represented as a random mixture of latent topics, where each topic is a probability distribution over words. The LDA graphical representation is shown is Fig. \ref{Fig:LDA}.

Let $\mathcal{D} = \{d_1, d_2, \ldots, d_{\mathcal{|D|}}\}$ is the corpus and $\mathcal{V} = \{w_1, w_2, \ldots, w_{\mathcal{|V|}}\}$ is the vocabulary of the corpus. A topic $z_j, 1 \leq j \leq K$ is represented as a multinomial probability distribution over the $|\mathcal{V}|$ words, $p(w_i|z_j), \sum_{i}^{|\mathcal{V}|}{p(w_i|z_j) = 1}$. LDA generates the words in a two-stage process: words are generated from topics and topics are generated by documents. More formally, the distribution of words given the document is calculated as follows:

\begin{equation}
	p(w_i|d) = \sum\limits_{j = 1}^K{p(w_i|z_j)p(z_j|d)}
\end{equation}

The LDA assumes the following generative process for the corpus $\mathcal{D}$:

\begin{enumerate}
	\item For each topic $k \in \{1, 2, \ldots, K\}$, sample a word distribution $\varphi_k \sim$ Dir($\beta$)
	\item For each document $d \in \{1, 2, \ldots, D \}$,
	\begin{itemize}
		\item[(a)] Sample a topic distribution $\theta_d$ $\sim$ Dir($\alpha$)
		\item[(b)] For each word $w_n$, where $n \in \{1, 2, \ldots, N\}$, in document $d$,
		\begin{itemize}
			\item[i.] Sample a topic $z_i$ $\sim$ Mult($\theta_d$)
			\item[ii.] Sample a word $w_n$ $\sim$ Mult($\varphi_{z_i}$)
		\end{itemize}
	\end{itemize}
\end{enumerate}

The joint distribution of the model (hidden and observed variables) is:

\begin{align}\nonumber
	&P(\phi_{1:K}, \theta_{1:\mathcal{D}}, z_{1:\mathcal{D}}, w_{1:\mathcal{D}}) = \prod_{j=1}^K{P(\phi_j|\beta)} \prod_{d=1}^{D}{P(\theta_d|\alpha)}\\& \times \left( \prod_{n=1}^N{P(z_{d,n}|\theta_d)P(w_{d,n}|\phi_{1:K},z_{d,n})} \right)
	\end{align}

\subsubsection{Inference and Parameter Estimation for LDA}
We now need to compute the \emph{posterior} distribution of the hidden variables, given the observed documents. Thus, the posterior is:

\begin{equation}
	P(\varphi_{1:K}, \theta_{1:D}, z_{1:D}| w_{1:D}) = \dfrac{P(\varphi_{1:K}, \theta_{1:D}, z_{1:D}, w_{1:D})}{P(w_{1:D})}
\end{equation}

This distribution is intractable to compute \cite{blei2003latent} due to the denominator (probability of seeing the observed corpus under any topic model).

While the posterior distribution (exact inference) is not tractable, a wide variety of approximate inference techniques can be used, including variational inference \cite{blei2003latent} and Gibbs sampling \cite{griffiths2004finding}. Gibbs sampling is a Markov Chain Monte Carlo \cite{gilks1996introducing} algorithm, trying to collect sample from the posterior to approximate it with an empirical distribution.

Gibbs sampling computes the posterior over topic assignments for every word as follows:

\begin{align}\nonumber
	&P(z_i=k|w_i = w, \textbf{z}_{-i}, \textbf{w}_{-i}, \alpha, \beta) =\\& \dfrac{n^{(d)}_{k,-i} + \alpha}{\sum_{k'=1}^K{n^{(d)}_{k',-i}} + K\alpha} \times
 		  \dfrac{ n^{(k)}_{w,-i} + \beta }{ \sum_{w'=1}^W{n^{(k)}_{w',-i}} + W \beta }
\end{align}

where $z_i = k$ is the topic assignment of word $i$ to topic $k$, $z_{-i}$ refers to the topic assignments of all other words. $n^{(k)}_{w,-i}$ is the number of times word $w$ assigned to topic $k$ excluding the current assignment. Similarly, $n^{(d)}_{k,-i}$ is the number of times topic $k$ is assigned to any words in document $d$ excluding the current assignment. For a theoretical overview on Gibbs sampling see \cite{carpenter2010integrating,heinrich2005parameter}.

LDA can be easily used as a module in more complicated models for more complex goals. Furthermore, LDA has been extensively used in a wide variety of domains. Chemudugunta et al. \cite{chemudugunta2008modeling} combined LDA with concept hierarchy to model documents. \cite{allahyari2015automatic,allahyari2016semantic} developed ontology-based topic models based on LDA for automatic topic labeling and semantic tagging, respectively. \cite{allahyari2016wise} proposed a knowledge-based topic model for context-aware recommendations. \cite{kataria2011entity,sen2012collective} defined more complex topic models based on LDA for entity disambiguation, \cite{allahyari2016discovering} and \cite{han2012entity} has proposed a entity-topic models for discovering coherence topics and entity linking, respectively. Additionally, many variations of LDA have been created such as supervised LDA (sLDA) \cite{blei2007supervised}, hierarchical LDA (hLDA) \cite{blei2003hierarchical} and Hierarchical pachinko allocation model (HPAM) \cite{mimno2007mixtures}.

\section{Information Extraction} 
Information extraction (IE) is the task of automatically extracting structured information from unstructured or semi-structured text. In other words information extraction can be considered as a limited form of full natural language understanding, where the information we are looking for are known beforehand \cite{hotho2005brief}. For example, consider the following sentence: ``\emph{Microsoft was founded by Bill Gates and Paul Allen on April 4, 1975.}''

We can identify following information:

\hspace{.5in}\texttt{FounderOf(\emph{\textrm{Bill Gates, Microsoft}})}

\hspace{.5in}\texttt{FounderOf(\emph{\textrm{Paul Allen, Microsoft}})}

\hspace{.5in}\texttt{FoundedIn(\emph{\textrm{Microsoft, April - 4 1975}})}\\

IE is one of the critical task in text mining and widely studied in different research communities such as information retrieval, natural language processing and Web mining. Similarly, It has vast application in domains such as biomedical text mining and business intelligence. See \cite{aggarwal2012mining} for some of the applications of information extraction.

Information extraction includes two fundamental tasks, namely, \emph{name entity recognition} and \emph{relation extraction}. The state of the art in both tasks are statistical learning methods. In the following we briefly explain two information extraction tasks.

\subsection{Named Entity Recognition (NER)}
A named entity is a sequence of words that identifies some real world entity, e.g. ``Google Inc'', ``United States'', ``Barack Obama''. The task of named entity recognition is to locate and classify named entities in free text into predefined categories such as \emph{person}, \emph{organization}, \emph{location}, etc. NER can not be completely done simply by doing string matching against a dictionary, because
\begin{inparaenum}[\itshape a\upshape)]
	\item dictionaries are usually incomplete and do not contain all forms of named entities of a given entity type.
	\item Named entities are frequently dependent on context, for example ``big apple'' can be the fruit, or the nickname of New York.
\end{inparaenum}

Named entity recognition is a preprocessing step in the relation extraction task and also has other applications such as in question answering \cite{aggarwal2012mining,li2002learning}. Most of the named entity recognition techniques are statistical learning methods such as hidden Markov models \cite{bikel1997nymble}, maximum entropy models \cite{Chieu:2003:NER:1119176.1119199}, support vector machines \cite{isozaki2002efficient} and conditional random fields \cite{settles2004biomedical}. 

\subsection{Hidden Markov Models}
Standard probabilistic classification techniques usually do not consider the predicted labels of the neighboring words. The probabilistic models which take this into account are hidden Markov model (HMM). Hidden Markov model assumes a Markov process in which generation of a label or an observation depends on one or a few previous labels or observations. Therefore, given a sequence of labels $\textbf{Y}=(y_1, y_2,\ldots, y_n)$ for an observation sequence $\textbf{X}=(x_1, x_2,\ldots, x_n)$, we have:
\begin{equation}
 y_i \sim p(y_i|y_{i-1}) \hspace{30pt} x_i \sim p(x_i|x_{i-1})
\end{equation}
Hidden Markov models have been successfully used in the named entity recognition task and speech recognition systems. For an overview on hidden Markov models see \cite{rabiner1989tutorial}.

\subsection{Conditional Random Fields}
Conditional random fields (CRFs) are probabilistic models for sequence labeling. CRFs first introduced by Lafferty et al. \cite{lafferty2001conditional}. We refer to the same definition of conditional random fields in \cite{lafferty2001conditional} on observations (data sequences to be labeled) and $\textbf{Y}$ (sequence of labels) as follows:\\

\textbf{\textsc{Definition}}. \emph{Let} $G=(V,E)$ \emph{be a graph such that} $\textbf{Y} = (\textbf{Y}_v)_{v \in V}$, \emph{so that} $\textbf{Y}$ \emph{is indexed by vertices of} $\textbf{G}$. \emph{Then} $(\textbf{X}, \textbf{Y})$ \emph{is a conditional random field, when the random variables} $\textbf{Y}_v$, \emph{conditioned on} $\textbf{X}$, \emph{obey Markov property with respect to graph, and}:

\begin{equation}
	p(\textbf{Y}_v|\textbf{X}, \textbf{Y}_w, w \neq v) = p(\textbf{Y}_v|\textbf{X}, \textbf{Y}_w, w \sim v)
\end{equation}

\emph{where} $w \sim v$ \emph{means} $w$ \emph{and} $v$ \emph{are neighbors in} $G$.

Conditional random fields are widely used in information extraction and part of speech tagging \cite{lafferty2001conditional}

\subsection{Relation Extraction}
Relation extraction is another fundamental information extraction task and is the task of seeking and locating the semantic relations between entities in text documents. There are many different techniques proposed for relation extraction. The most common method is consider the task as a classification problem: Given a couple of entities co-occurring in a sentence, how to categorize the relation between two entities into one of the fixed relation types. There is a possibility that relation span across multiple sentences, but such cases are rare, thus, most of existing work have focused on the relation extraction within the sentence. Many studies using the classification approach for relation extraction have been done such as \cite{chan2010exploiting,chan2011exploiting,jiang2007systematic,kambhatla2004combining,guodong2005exploring}. 

\section[Biomedical Ontologies and Text Mining]{Biomedical Ontologies and Text Mining for Biomedicine and Healthcare}
One of the domains where text mining is tremendously used is biomedical sciences. Biomedical literature is growing exponentially, Cohen and Hunter \cite{cohen2008getting} show that the growth in PubMed/MEDLINE publications is phenomenal, which makes it quite difficult for biomedical scientists to assimilate new publications and keep up with relevant publications in their own research area. 

In order to overcome this text information overload and transform the text into machine-understandable knowledge, automated text processing methods are required. Thus, text mining techniques along with statistical machine learning algorithms are widely used in biomedical domain. Text mining methods have been utilized in a variety of  biomedical domains such as protein structure prediction, gene clustering, biomedical hypothesis and clinical diagnosis, to name a few. In this section, we briefly describe some of the relevant research in biomedical domain, including biomedical ontologies and then proceed to explain some of the text mining techniques in biomedical discipline applied for basic tasks of named entity recognition and relation extraction. 

\subsection{Biomedical Ontologies}
We first define the concept of \emph{ontology}. We use the definition presented in W3C's OWL Use Case and Requirements Documents\footnote{\texttt{http://www.w3.org/TR/webont-req/}} as follows:\\

\emph{An ontology formally defines a common set of terms that are used to describe and represent a domain. An ontology defines the terms used to describe and represent an area of knowledge}.\\

According to the definition above, we should mention a few points about ontology: 
\begin{inparaenum}[\itshape 1\upshape)]
	\item Ontology is domain specific, i.e., it is used to describe and represent \emph{an area of knowledge} such as area in education, medicine, etc \cite{yu2011developer}. 
	\item Ontology consists of terms and relationships among these terms. Terms are often called \emph{classes} or concepts and relationships are called \emph{properties}.\\

\end{inparaenum}

There are a great deal of biomedical ontologies. For a comprehensive list of biomedical ontologies see Open Biomedical Ontologies (OBO)\footnote{\texttt{http://www.obofoundry.org/}} and the National Center for Biomedical Ontology (NCBO)\footnote{\texttt{http://www.bioontology.org/}}. The NCBO ontologies are accessed and shared through BioPortal\footnote{\texttt{http://bioportal.bioontology.org/}}. In the following, we briefly describe one the most extensively-used ontologies in  biomedical domain:\\

\textbf{Unified Medical Language System (UMLS):} UMLS\footnote{\texttt{https://uts.nlm.nih.gov/home.html}} \cite{mc1993unified} is the most comprehensive knowledge resource, unifying over 100 dictionaries, terminologies and ontologies in its Metathesaurus (large vocabulary whose data are collected from various biomedical thesauri) which is designed and maintained by National Library of Medicine (NLM). It provides a mechanism for integrating \cite{bodenreider2004unified} all main biomedical vocabularies such as MeSH, Systematized Nomenclature of Medicine Clinical Terms (SNOMED CT), Gene Ontology (GO), etc. 

It also provides a semantic network that explains the relations between Metathesaurus entries, i.e., a dictionary that includes lexicographic information about biomedical terms and common English words and a set of lexical tools. Semantic network contains semantic types and semantic relations. Semantic types are categories of Metathesaurus entries (concepts) and semantic relations are relationships between semantic types. For more information about UMLS, see \cite{yoo2008biomedical,bodenreider2004unified}.

Apart from the aforementioned ontologies and knowledge sources, there are various ontologies more specifically focused on biomedical sub-domains. For example, the Pharmacogenomics Knowledge Base \footnote{\texttt{http://www.pharmgkb.org/}}, consists of clinical information including dosing guidelines and drug labels, potentially clinically actionable gene-drug associations and genotype-phenotype relationships.

The ontologies and knowledge bases described earlier are extensively used by different text mining techniques such as information extraction and clustering in the  biomedical domain.

\subsection{Information Extraction}
As mentioned before (section 5), \emph{information extraction} is the task of extracting structured information from unstructured text in a automatic fashion. In biomedical domain, unstructured text comprises mostly scientific articles in biomedical literature and clinical information found in clinical information systems. Information extraction is typically considered as a preprocessing step in other biomedical text mining applications such as question answering \cite{athenikos2010biomedical}, knowledge extraction \cite{savova2010mayo, 2017arXiv170607992T}, hypothesis generation\cite{liekens2011biograph, cohen2005survey} and summarization \cite{hersh2008information}. 

\subsubsection{Named Entity Recognition (NER)}  
Named Entity Recognition is the task of information extraction which is used to locate and classify biomedical entities into categories such as protein names, gene names or diseases \cite{leser2005makes}. Ontologies can be utilized to give semantic, unambiguous representation to extracted entities. NER are quite challenging in biomedical domain, because: 
\begin{enumerate}
	\item There is a huge amount of semantically related entities in biomedical domain and is increasing quickly with the new discoveries done in this field. This non-stop growing of the volume of entities is problematic for the NER systems, since they depends on dictionaries of terms which can never be complete due to continues progress in scientific literature.

	\item In biomedical domain, the same concept may have many different names (synonyms). For example, ``heart attack'' and ``myocardial infarction'' point to the same concept and NER systems should be able to recognize the same concept regardless of being expressed differently.

	\item Using acronyms and abbreviations is very common in biomedical literature which makes it complicated to identify the concepts these terms express.
\end{enumerate}

It is critical for NER systems to have high quality and perform well when analyzing vast amounts of text. Precision, recall and \textit{F}-score are typical evaluation methods used in NER systems. Even though there are some challenges to obtain and compare evaluation methods reliably, e.g. how to define the boundaries of correctly identified entities, NER systems have demonstrated to achieve good results in general.

NER methods are usually grouped into several different approaches:


\begin{itemize}
	\item \textbf{Dictionary-based approach}, is one of the main biomedical NER methods which uses a exhaustive dictionary of biomedical terms to locate entity mentions in text. It decides whether a word or phrase from the text matches with some biomedical entity in the dictionary. Dictinary-based methods are mostly used with more advanced NER systems.
	
	\item \textbf{Rule-based approach}, defines rules that specifies patterns of biomedical entities. Gaizauskas et at. \cite{gaizauskas2003protein} have used context free grammars to recognize protein structure.

	\item \textbf{Statistical approaches}, basically utilize some machine learning methods typically supervised or semi-supervised algorithms \cite{usami2011automatic} to identify biomedical entities. Statistical methods are often categorized into two different groups:

	\begin{enumerate}
		\item \textbf{Classification-based approaches}, convert the NER task into a classification problem, which is applicable to either words or phrases. Naive Bayes \cite{nobata1999automatic} and Support Vector Machines \cite{kazama2002tuning,takeuchi2005bio,mitsumori2005gene} are among the common classifiers used for biomedical NER task.

		\item \textbf{Sequence-based methods}, use complete sequence of words instead of only single words or phrases. They try to predict the most likely tag for a sequence of words after being trained on a training set. Hidden Markov Model (HMM) \cite{collier2000extracting,shen2003effective,zhao2004named}, Maximum Entropy Markov Model \cite{corbett2008cascaded} and Conditional Random Fields (CRF) \cite{settles2004biomedical} are the most common sequence-based approaches and CRFs have frequently demonstrated to be better statistical biomedical NER systems.

		\item \textbf{Hybrid methods}, which rely on multiple approaches, such as combining dictionary- or rule-based techniques with statistical methods. \cite{sasaki2008make} introduced a hybrid method in which they have use a dictionary-based method to locate known protein names along with a part-of-speech tagging (CRF-based method).
	\end{enumerate}
\end{itemize}

\subsubsection{Relation Extraction}
Relation extraction in Biomedical domain involves determining the relationships among biomedical entities. Given two entities, we aim at locating the occurrence of a specific relationship type between them. The associations between entities are usually binary, however, it can include more than two entities. In the genomic area, for example, the focus is mostly on extracting interactions between genes and proteins, such as protein-protein or gene-diseases relationships. Relation extraction comes across similar challenges as NER, such as creation of high quality annotated data for training and assessing the performance of relation extraction systems. For more information see \cite{ananiadou2010event}.

There are many different approaches for biomedical relation extraction. The most straightforward technique is based on where the entities co-occur. If they mentioned together frequently, there is high chance that they might be related in some way. But we can not recognize the type and direction of the relations by using only the statistics. Co-occurrence approaches are usually give high recall and low precision.

Rule-based approaches are another set of methods used for biomedical relation extraction. Rules can be defined either manually by domain experts or automatically obtained by using machine learning techniques from an annotated corpus. Classification-based approaches are also very common methods for relation extractions in biomedical domain. There is a great work done using supervised machine learning algorithms that detects and discovers various types of relations, such as \cite{bundschus2008extraction} where they identify and classify relations between diseases and treatments extracted from PubMed abstracts and between genes and diseases in human GeneRIF database.

\subsection{Summarization}
One of the common biomedical text mining task which largely utilizes information extraction tasks is \emph{summarization}. Summarization is the task of identifying the significant aspects of one or more documents and represent them in a coherent fashion \emph{automatically} . it has recently gained a great attention because of the huge growth of unstructured information in biomedical domain such as scientific articles and clinical information.

Biomedical summarization is often application oriented and may be applied for different purposes. Based on their purpose, a variety of document summaries can be created such as single-document summaries which targets the content of individual documents and multi-document summaries where information contents of multiple documents are considered.

The evaluation of summarization methods is really challenging in biomedical domain. Because deciding whether or not a summary is ``good'' is often subjective and also manual evaluations of summaries are laborious to carry out. There is a popular automatic evaluation technique for summaries that is called ROUGE (Recall-Oriented Understudy for Gisting Evaluation). ROUGE measures the quality of an automatically produced summary by comparing it with ideal summaries created by humans. The measure is calculated by counting the overlapping words between the computer-generated summary and the ideal human-produced summaries. For a comprehensive overview of various biomedical summarization techniques, see \cite{aggarwal2012mining}.

\subsection{Question Answering}
Question answering is another biomedical text mining task where significantly exploits information extraction methods. \emph{Question answering} is defined as the process of producing accurate answers to questions posed by humans in a natural language. Question answering is very critical in biomedical domain due to data overload and constant growth of information in this field. 

In order to generate precise responses, question answering systems make extensive use of natural language processing techniques. The primary steps of question answering system is as follows: 
\begin{inparaenum}[\itshape a\upshape)]
	\item The system receives a natural language text as input.
	\item Using linguistic analysis and question categorization algorithms, the system determines the type of the posed question and the answer it should produce.
	\item Then it generates a query and passes it to the document processing phase.
	\item In the document processing phase, system feeds the query, system sends the query to a search engine, gets back the retrieved documents and extracts the relevant snippets of text as candidate answers, and send them to answering processing stage.
	\item Answering processing stage, analyzes the candidate answers and ranks them according to the degree they match the expected answer type that was established in the question processing step.
	\item The top-ranked answer is selected as the output of the question answering system.
\end{inparaenum}	 

Question answering systems in biomedical discipline have recently begun to utilize and incorporate semantic knowledge throughout their processing steps to create more accurate responses. These biomedical \emph{semantic knowledge-based} systems use various semantic components such as semantic meta-data represented in knowledge sources and ontologies and semantic relationships to produce answers. See \cite{aggarwal2012mining,athenikos2010biomedical} for a complete overview of different biomedical question answering techniques. 

\section{Discussion}
In this article we attempted to give a brief introduction to the field of text mining. We provided an overview of some of the most fundamental algorithms and techniques which are extensively used in the text domain. This paper also overviewed some of important text mining approaches in the biomedical domain. Even though, it is impossible to describe all different methods and algorithms throughly regarding the limits of this article, it should give a rough overview of current progresses in the field of text mining.

Text mining is essential to scientific research given the very high volume of scientific literature being produced every year \cite{gutierrez2015within}. These large archives of online scientific articles are growing significantly as a great deal of new articles are added in a daily basis. While this growth has enabled researchers to easily access more scientific information, it has also made it quite difficult for them to identify articles more pertinent to their interests. Thus, processing and mining this massive amount of text is of great interest to researchers. 

\begin{acks}
This project was funded in part by Federal funds from the US National Institute of Allergy and Infectious Diseases, National Institutes of Health, Department of Health and Human Services under contract \#HHSN272201200031C, which supports the Malaria Host-Pathogen Interaction Center (MaHPIC).
\end{acks}

\section{Conflict of Interest}
The author(s) declare(s) that there is no conflict of interest regarding the publication of this article.

\bibliography{compbib}

\bibliographystyle{ACM-Reference-Format}

\end{document}